\pdfoutput=1
\documentclass[runningheads]{llncs}
\usepackage[T1]{fontenc}
\usepackage{graphicx}
\usepackage{booktabs}
\usepackage{multirow}
\usepackage{tabularx}
\usepackage{amsmath,amssymb}
\usepackage{xcolor}
\usepackage[hidelinks]{hyperref}
\usepackage[capitalise,noabbrev]{cleveref}
\definecolor{pumaRed}{HTML}{B1243C}
\crefname{figure}{Fig.}{Figs.}
\crefname{table}{Table}{Tables}
\crefname{section}{Section}{Sections}
\begin{document}

\title{Risk Control Certifies Rates, Not Magnitudes: An Audit of Selective Fetal Biometry}
\titlerunning{Risk Control Certifies Rates, Not Magnitudes}
\author{Zhengshuai Yang\inst{1} \and
Linna Jiang\inst{1} \and
Changyun Li\inst{1}\fnmsep\thanks{Corresponding authors.} \and
Guihu Zhao\inst{2,\star}}
\authorrunning{Z. Yang et al.}
\institute{College of Information and Intelligent Science and Technology, Hunan Agricultural University, Changsha 410128, China\\
\email{yhvguk@stu.hunau.edu.cn}, \email{362330193@qq.com} \and
National Clinical Research Center for Geriatric Disorders, Xiangya Hospital, Central South University, Changsha 410008, China\\
\email{ghzhao@csu.edu.cn}}

\maketitle
\begin{abstract}
Prenatal ultrasound biometry must return a head circumference (HC) and refuse when the image cannot support one. Selective prediction trades coverage against error, but its finite-sample guarantees bound the \emph{rate} of gross errors, not their magnitude. We audit a tool-grounded biometry pipeline on two separately pre-registered four-way splits of the 999 labelled HC18 images, retraining the segmenter three times on each. Calibrating the contour-completeness threshold by exact Clopper--Pearson risk control returns a certificate indistinguishable from a hand-set floor, yet the pipeline still emits a 271.3\,mm error: a small, geometrically complete ring fitted to an image whose true HC is 324.6\,mm. The accepted candidate scores 0.94 completeness, so the score cannot see the failure, and one outlier among $\approx$150 answers leaves the certified risk bound ($\alpha{=}0.10$) intact. A second, untouched split reproduces the mechanism on a different image (233.3\,mm). Constraining the \emph{admissibility region}---the anatomical range in which the measurand is defined---removes the family instead: raising the lower bound from 50 to 55\,mm costs 0.4 coverage points and cuts the worst-case error on both splits to 11.4 and 11.1\,mm, whereas suppressing it through the completeness threshold costs 19.0 and 12.1 points. At an anatomically defined 70\,mm the frozen tests give 1.91 and 1.81\,mm MAE.
\keywords{selective prediction \and distribution-free risk control \and fetal ultrasound biometry \and head circumference \and failure-mode audit.}
\end{abstract}

\section{Introduction}
\label{sec:intro}

An assistant that reports fetal head circumference (HC) from ultrasound must return a physically scaled measurement and refuse when the image cannot support one. HC, a routine indicator of fetal growth, can be estimated from a skull contour and the pixel spacing~\cite{vandenheuvel2018data}; inside a vision--language model (VLM) the harder question is not how to compute the value but when it can be trusted.

Producing a reliable numeric measurement and deciding whether to trust it are distinct challenges that expose the limitations of current VLMs. A general end-to-end VLM often struggles with precise geometric biometry; it may omit a measurement, hallucinate a scale from pure visual features, or blindly accept an incorrect external observation. Conversely, a purely geometric tool lacks semantic understanding and can return a mathematically plausible ellipse fitted to the wrong anatomical structure. Consequently, evaluating only measurement error on a selected high-quality subset does not establish reliable biometry across a full clinical cohort.

We study PUMA (Prenatal Ultrasound Multimodal Assistant), a tool-grounded VQA system that separates geometric computation, candidate selection, and anatomical verification into auditable steps. PUMA computes the measurement before generating an answer: a contour segmenter and an ellipse-fitting tool compute HC in millimetres, a completeness gate filters partial candidates, additional views are considered if the initial view fails, and the VLM verifies the accepted observation or abstains. This design exposes an interpretable selection signal, allowing us to evaluate measurement error and the proportion of answered questions---and, as we show, to measure how far a calibrated selection rule can actually be trusted.

In summary, our main contributions are:
1) We audit PUMA, a tool-grounded biometry pipeline, on a curated prenatal benchmark (\cref{sec:data,sec:benchmark}): against 1/96 HC answers for a direct VLM, the delivered configuration answers 66/96 at 1.76\,mm MAE, and the paired benchmark with an ablation shows that the geometric gate, not the language model, is the decisive component.
2) Across two pre-registered four-way splits with three retrained seeds each, exact Clopper--Pearson calibration of the completeness threshold is indistinguishable from a hand-set floor---and remains compatible with a 271.3\,mm error, because at $n\approx150$ a single outlier barely moves a gross-error rate (\cref{sec:cohort}).
3) We show that calibrating the anatomical admissibility region, rather than the score threshold, removes this family for 0.4 coverage points instead of 19, and confirm it with one-shot frozen-test evaluations on both splits (\cref{sec:cohort}).

\section{Related Work}
\label{sec:related}

\emph{Fetal biometry.} HC18 supports automated head measurement~\cite{vandenheuvel2018data}, including studies of human agreement~\cite{sinclair2018human} and contour estimation~\cite{chen2022efficienthc}, while plane detection and verification~\cite{baumgartner2017sononet,yaqub2017neurosono} complement it with FETAL\_PLANES\_DB~\cite{burgos2020fetalplanes}. We study acceptance of tool outputs, without claiming superiority over dedicated biometry systems.

\emph{Medical VQA and tools.} VQA-RAD, SLAKE and PMC-VQA support biomedical question answering~\cite{lau2018vqarad,liu2021slake,zhang2023pmcvqa} and multimodal assistants~\cite{chun2023llavamed,moor2023medflamingo}. Retrieval supplies textual context~\cite{lewis2020rag}; biometry additionally requires geometry and scale. Unlike model-directed tool invocation~\cite{schick2023toolformer,yao2023react}, PUMA measures before generating and then requests verification.

\emph{Selective prediction.} Reject-option methods trade answer coverage against error~\cite{chow1970reject,geifman2017selective,geifman2019selectivenet}. Calibration~\cite{guo2017calibration} and distribution-free risk control~\cite{angelopoulos2023gentle,bates2021distribution} formalise this tradeoff; we apply exact Clopper--Pearson bounds on a reserved calibration split.

\section{Data: Curation and Split Audit}
\label{sec:data}

The catalogue holds 98{,}607 records across ten adapters, from HC18 (999 labelled images), FETAL\_PLANES\_DB, PMC-VQA, SLAKE, VQA-RAD and three obstetric text archives; MedDialog (17{,}152) is quarantined because its reuse terms are unverified. Connected groups link records by patient, examination or article identifiers, exact image hashes and normalised-text duplicates, and official test assignments take precedence. Only 126 native prenatal questions survive domain screening, and 82 PMC-VQA figures pass a caption rule requiring ultrasound terminology and prenatal context. HC18 has no patient identifiers, so its isolation is image-level.

\section{Method}
\label{sec:method}

\subsection{Answer Contract and Strategies}
\label{sec:contract}

The benchmark shares one served generator (\texttt{qwen3.8-flash}, snapshot 2026-09-02), temperature 0, a 1{,}200-token budget and a 240\,s timeout. Protocol \texttt{puma-v1.4} returns \texttt{answer}, \texttt{abstained}, \texttt{limitations} and \texttt{evidence\_ids}: a short label, at most ten words, or a value in mm, with qualifications separate. The original benchmark counts empty successful responses as abstentions; later logs distinguish failed generation from refusal.

\texttt{direct\_vqa} uses image and question; \texttt{rag\_vqa} adds retrieval context; \texttt{staged\_agent} allocates budgeted time to observation, answering, and verification; \texttt{tool\_vqa} adds measurement and requests anatomical verification.

\subsection{Deterministic Biometry Tool}
\label{sec:tool}

\begin{figure*}[t]
\centering
\includegraphics[width=0.82\linewidth]{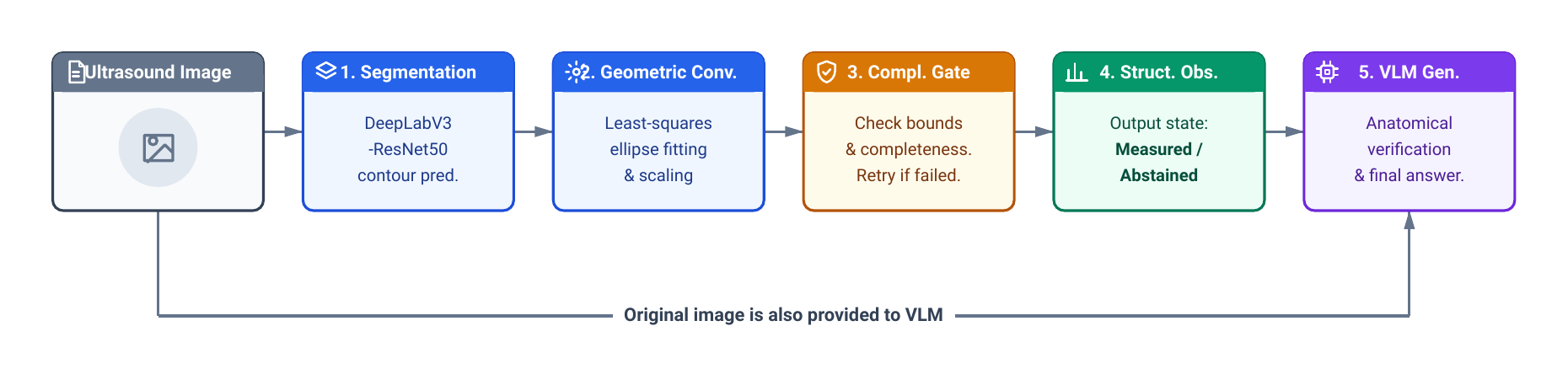}
\caption{Measurement runs before VLM generation. The original end-to-end runs use $L=25$\,mm; offline audits and the second-family run also examine $L=50$\,mm. The VLM may accept the observation or abstain.}
\label{fig:pipeline}
\end{figure*}

The measurement pipeline in \cref{fig:pipeline} separates contour estimation, physical conversion, and candidate selection before the VLM receives an observation. By isolating these steps, PUMA provides an auditable numeric measurement rather than relying on the VLM to estimate geometry.

\emph{Segmentation.} To ground the final answer in physical pixels, a DeepLabV3--ResNet50 model~\cite{chen2017deeplabv3} predicts skull contours. Images are non-uniformly resized to $512\times512$ pixels, and predictions are mapped back to source coordinates. The network is trained on 736 HC18 images with physician masks using BCE plus Dice loss. The delivered v2 recipe further adds shadow-sector occlusion and boundary-band BCE to improve robustness on poor-quality scans. Any numeric answer therefore originates from a concrete image region rather than a VLM estimate.

\emph{Geometric Conversion.} Converting a raw pixel contour to a continuous circumference measurement requires geometric fitting. After morphological closing, connected-component rims are fitted using a direct least-squares conic method~\cite{fitzgibbon1999ellipse}. The HC is computed as the Ramanujan ellipse perimeter multiplied by the DICOM pixel spacing. We also define contour completeness $c$ as the fraction of 16 angular bins around the fitted centre that are occupied by rim pixels. By computing this deterministic geometric metric, the system can interpret the quality of the observation independently of the VLM's semantic understanding.

\emph{Completeness Gate and Additional Views.} A VLM might blindly accept a mathematically plausible but anatomically incorrect ellipse (e.g., fitted to a small artifact). To prevent this, a geometric gate filters candidates before VLM verification. A candidate is eligible only if its completeness $c \geq \tau$ and its HC lies in $[L, 450]$\,mm (the original runs use $\tau=0.7$ and $L=25$\,mm). If the initial view yields no eligible component, the tool acts on four shifted and two zoomed views, reversing the zoom scale on the output perimeter. This hard gate prevents the VLM from answering when no valid contour exists, thereby lowering the risk of gross errors.

\emph{Structured Observations.} Finally, the tool exports an observation state (\texttt{measured}, \texttt{abstained}, \texttt{error}, or \texttt{unavailable}) to the VLM. On accepted HC answers, the VLM copies the reported value exactly. This structured handoff reduces the VLM's role to verifying whether the proposed measurement semantically aligns with the question; the value is passed through rather than estimated.

\section{Experimental Setup}
\label{sec:setup}

The development benchmark has 200 validation questions over 163 groups: 97 plane, 96 HC and seven native prenatal questions, selected by stratified group round-robin. Brain sub-planes were excluded after a pilot. Interleaved strategies produced 800 answers without call errors; two retrieval usage-censored rows are excluded from the table.

We report exact match and token-F1 with group-bootstrap 95\% CIs. MAE and median absolute error use returned measurements; answer rate uses all questions. Gross HC error means absolute error $\geq7$\,mm, with rate among returned readings. The 2\,mm hit threshold and $\pm2\%$ rule are engineering criteria, not clinical tolerances.

A rebuilt three-judge panel reproduces 54/66 numeric acceptances (81.8\%). Across 80 audited non-numeric responses, acceptance is 18/80 (22.5\%) and Fleiss $\kappa=0.931$.

\section{Results}
\label{sec:results}

\subsection{Paired Strategy Benchmark}
\label{sec:benchmark}

\begin{table}[t]
\centering
\caption{Plane accuracy and HC metrics are conditional on answered questions; tokens are run totals in thousands and latency the run-record mean. Only the first four rows share one run, so only those support the paired tests; the delivered run holds the tool arm alone and its plane accuracy is unpaired. $^\star$HC biometry only. Offline replay of the same candidates with the $L{=}50$\,mm floor gives 66 answers at 1.78\,mm, so the language model adds no measurement.}
\label{tab:benchmark}
\footnotesize
\setlength{\tabcolsep}{3.5pt}
\begin{tabular}{@{}lcccccccc@{}}
\toprule
\multirow{2}{*}{Strategy} & \multicolumn{2}{c}{Plane QA ($n{=}97$)} & \multicolumn{4}{c}{HC Biometry ($n{=}96$)} & \multirow{2}{*}{\shortstack{Tok.\\(k)}} & \multirow{2}{*}{\shortstack{Lat.\\(s)}} \\
\cmidrule(lr){2-3} \cmidrule(lr){4-7}
 & Ans. & Acc. & Ans. & MAE & Med. & $>$7\,mm & & \\
\midrule
\multicolumn{9}{@{}l}{\emph{Paired benchmark: one interleaved four-arm run (\texttt{89d79bf1})}}\\
Direct & 88 & .841 & \phantom{0}1 & 41.0 & 41.0 & 1 & 290 & \phantom{0}9.3 \\
Retrieval & 92 & .880 & \phantom{0}9 & 12.7 & 13.3 & 6 & 432 & 17.9 \\
Staged & 77 & .766 & \phantom{0}0 & --- & --- & 0 & 1{,}424 & 43.0 \\
Tool VQA & 93 & .871 & 71 & 6.50 & 2.30 & 16 & 373 & 14.8 \\
\midrule
\multicolumn{9}{@{}l}{\emph{Delivered run (\texttt{98364160}), tool arm only, not paired with the block above}}\\
Tool VQA + floor$^\star$ & 93 & .914 & 66 & \phantom{0}\textbf{1.76} & \phantom{0}\textbf{1.19} & \textbf{1} & 318 & 13.0 \\
\midrule
\multicolumn{9}{@{}l}{\emph{Deterministic replay of the same stored candidates (no language model)}}\\
Tool+Gate, $L{=}25$\,mm & --- & --- & 67 & 5.74 & 1.19 & 2 & \phantom{0}\phantom{0}0 & \phantom{0}1.2 \\
Tool+Gate, $L{=}50$\,mm & --- & --- & 66 & 1.78 & 1.18 & 1 & \phantom{0}\phantom{0}0 & \phantom{0}1.2 \\
\bottomrule
\end{tabular}\\
\end{table}

Tool grounding is what makes HC answers possible, and the corrected floor is what makes them accurate (\cref{tab:benchmark}). On 97 plane questions it holds 87.1\% accuracy among 93 answered, matching retrieval's 88.0\%. On the 96 HC questions the direct VLM abstains on 95/96 (its single attempt off by 41.0\,mm) and retrieval answers nine at 12.7\,mm MAE, whereas tool grounding answers 71 at 6.50\,mm; the delivered run with the corrected floor answers 66 at 1.76\,mm with one gross error. 

Evaluating the deterministic tool without VLM verification yields 67 answers at 5.74\,mm MAE because of one degenerate 28.1\,mm ring; adding the 50\,mm floor gives 66 answers at 1.78\,mm, matching the delivered run's 66 at 1.76\,mm. Within this benchmark the language model therefore changes nothing that the geometric gate does not already change---an ablation we revisit in \cref{sec:cohort}.

A second served family (\texttt{gpt-6}, snapshot 2026-09-12) reproduces the pattern: 64 HC readings at 1.80\,mm MAE, with 14\% abstention against 43\%; 26 of 400 calls fail and are excluded.

\subsection{Completeness Gating and Additional Views}
\label{sec:gate}

With v2 weights, free VLM checking answers 77/96 at 6.49\,mm MAE with 22 gross errors; gating gives 48 at 1.80\,mm with one and the six-view fallback 66 at 1.76\,mm with one. Fourteen of sixteen v1 gross errors involve incomplete contours ($c=0.50$--$0.62$), while the VLM rejects one high-completeness false contour worth 266.6\,mm. These runs use the original 25\,mm floor.

\subsection{Admissibility Audit on Two Pre-Registered Splits}
\label{sec:cohort}

A whole-cohort replay already shows which family the pipeline fails on: among 999 labelled images, 45 readings below 50\,mm with completeness $c\geq0.7$ account for 44 of 53 gross errors. We therefore audit twice whether raising the floor removes them under a clean, one-shot evaluation. Each four-way partition fixes a 58/12/15/15 ratio (train/model selection/calibration/frozen test): calibration sets are $n{=}177$ (split 1) and $n{=}149$ (split 2). To prevent data leakage, split 2 was pre-registered with candidate floors $L \in \{50, 55, 70\}$\,mm fixed before its frozen test was evaluated. We evaluate $\tau \in [0.50, 1.00]$ using exact Clopper--Pearson binomial bounds with Bonferroni grid correction rather than asymptotic bounds, because at $n\approx150$ Hoeffding is excessively conservative ($0/105$ yields $0.105$ vs exact $0.022$).

\begin{table}[t]
\centering
\caption{Frozen-test evaluation on two pre-registered splits, $\tau{=}0.70$ throughout, mean $\pm$ sd over three retrained seeds; worst case is the largest absolute error in the split. The second split was registered after the first failure was identified, so it is a confirmatory replication.}
\label{tab:cohort}
\footnotesize
\setlength{\tabcolsep}{4pt}
\begin{tabular}{@{}lccccc@{}}
\toprule
Split / floor $L$ & Cov. & $n$ & MAE (mm) & Gross ($\geq$7\,mm) & Worst case (mm) \\
\midrule
\multicolumn{6}{@{}l}{\emph{Split 1, frozen test ($n{=}156$)}}\\
$L{=}50$\,mm & .722 & 112.7 & \phantom{0}3.51\,$\pm$\,1.42 & 2.39\% & 271.3 \\
$L{=}55$\,mm & .718 & 112.0 & \textbf{1.93\,$\pm$\,0.03} & 1.82\% & \phantom{0}\textbf{11.4} \\
\midrule
\multicolumn{6}{@{}l}{\emph{Split 2, frozen test ($n{=}152$)}}\\
$L{=}50$\,mm & .750 & 114.0 & \phantom{0}2.53\,$\pm$\,1.13 & 1.41\% & 233.3 \\
$L{=}55$\,mm & .748 & 113.7 & \textbf{1.81\,$\pm$\,0.17} & 1.10\% & \phantom{0}\textbf{11.1} \\
\bottomrule
\end{tabular}
\end{table}

\begin{figure*}[t]
\centering
\includegraphics[width=0.72\linewidth]{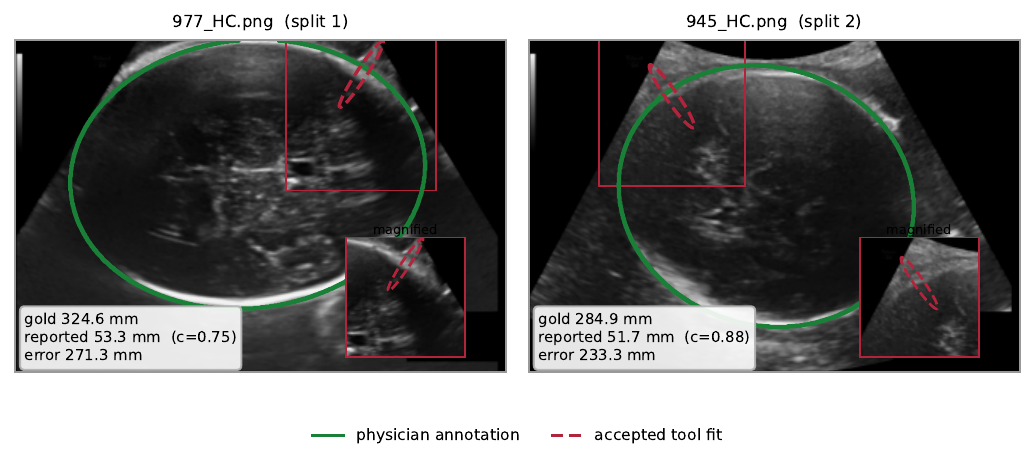}
\caption{The two failures that passed the completeness gate. Green: physician annotation; red dashed: the accepted conic fit, magnified in the inset. \emph{Left} (split 1): \texttt{977\_HC.png}, gold 324.6\,mm, fit 53.3\,mm at $c{=}0.75$, error 271.3\,mm. \emph{Right} (split 2): \texttt{945\_HC.png}, gold 284.9\,mm, fit 51.7\,mm at $c{=}0.88$, error 233.3\,mm. Both fits sit on the bright specular arc at the skull boundary, which is smooth and therefore scores high completeness.}
\label{fig:failures}
\end{figure*}

\emph{A reproducible failure mode.} At $L{=}50$ the pipeline emits one error two orders of magnitude larger than the median in both splits, from two different images (\cref{fig:failures}): \texttt{977\_HC.png} (gold 324.6\,mm, reported 53\,mm; error 271.3\,mm) and \texttt{945\_HC.png} (gold 284.9\,mm, reported 52\,mm; error 233.3\,mm). Across the six retrained models the accepted candidates for these two images are reported at 51.6--54.9\,mm with completeness 0.75--0.94, so a gate built on $c$ admits them and the 50\,mm floor sits just below their band.

\emph{Risk control certifies rates, not magnitudes.} Calibrating $\tau$ by exact Clopper--Pearson bounds at $\alpha{=}0.10$ returns $\tau^*{=}0.63$--$0.69$ on split 1, below the hand-set 0.70; the calibrated and fixed operating points are then identical to the reported precision in two of three seeds, and MAE moves from 3.51 to 3.53\,mm. On split 2 the same procedure selects $L{=}50$ in all three seeds, and all six calibration partitions certify it at $\alpha{=}0.10$ (upper bounds 0.070--0.097)---including the two that already contain an error above 250\,mm. The frozen test then still returns the 233.3\,mm error. The certificate is not wrong: with $n\approx150$ answers, a single gross error leaves the upper bound on the gross-error rate below 10\%. The guarantee constrains how \emph{often} the pipeline is wrong, not how \emph{badly}, so a certified operating point can still be catastrophically wrong once.

\emph{Constraining the admissibility region.} Because this failure is defined by where the measurement is physically possible, the effective parameter is the admissibility region $[L,450]$\,mm---the anatomical range in which the measurand is defined---not the score threshold. Raising $L$ from 50 to 55\,mm costs 0.4 coverage points and reduces the worst-case error from 271.3 to 11.4\,mm on split 1 and from 233.3 to 11.1\,mm on split 2. Suppressing the same three errors through $\tau$ instead requires $\tau\geq0.94$, the extreme of the score range, and costs 19.0 coverage points on the split-1 frozen tests and 12.1 on split 2. With the anatomically defined $L{=}70$\,mm---below the range in which the trans-thalamic plane is used---the split-2 calibration partitions certify at $\alpha{=}0.10$ for all three seeds (upper bounds 0.099, 0.057, 0.085) and the frozen tests give 1.91\,$\pm$\,0.02\,mm and 1.81\,$\pm$\,0.17\,mm MAE at 69.4\% and 73.9\% coverage.

\subsection{Secondary Checks}
\label{sec:transfer}

\emph{Secondary checks.} The segmenter reaches a mask-level contour MAE of 9.49\,mm (median 3.03\,mm) on the validation partition, against 10.31\,mm for the v1 recipe; these are training-diagnostic errors, and the end-to-end biometry MAE above is the system's operating performance. For context, human inter-observer difference on HC18 averages 2--3\% ($\approx$2--5\,mm across gestations)~\cite{sinclair2018human,vandenheuvel2018data}, placing the delivered 1.76\,mm MAE within expert variability while masking catastrophic gross failures. Forced to answer, the second family returns 31 numeric readings on the 96 HC questions at 4.18\,mm MAE with six gross errors (19.4\%); on 200 external plane questions 167 calls return at exact match 0.72 without abstaining.

\section{Discussion and Limitations}
\label{sec:discussion}

\emph{What the audit establishes.} The main result concerns the shape of the guarantee rather than the pipeline: selecting on a score with finite-sample risk control bounds how \emph{often} the system is grossly wrong, not how badly. At the sample sizes a calibration partition affords, an operating point can be certified at $\alpha{=}0.10$ and still return a 233\,mm error, and threshold calibration cannot remove that possibility, because the offending candidate is geometrically complete. Against a direct VLM, tool grounding is what makes millimetre answers possible at all, and the geometric gate rather than the language model suppresses degenerate candidates.

\emph{Practical consequence.} The parameter that matters is the admissibility region---the range in which the measurand is physically defined---and it should come from anatomy rather than from tuning. Here the admissibility bound removes the failure family for 0.4 coverage points, while the score threshold needs 0.94 and roughly twenty times as many. Such bounds are cheap to state and transfer to any target whose plausible size is known in advance; score thresholds are not.

\emph{Limitations.} HC18 carries no patient identifier, so both splits are image-level and cannot exclude perceptual near-duplicates. The second split was registered after the failure was identified on the first, and the $L$ values tested on it were fixed before its evaluation, so it tests a pre-specified hypothesis rather than searching for one. The measurement is deterministic by design: the language model carries the answer contract and the refusal decision, not the number, which is why gating alone reproduces the delivered MAE. Assessment is retrospective and automated rather than clinical, and the served families are compared as deployed rather than at matched weights. On 335 unlabelled official-test images the single-view answer rate is 0.573, matching 0.573 on labelled images, so the operating behaviour is stable across the two cohorts.

\section{Conclusion}
\label{sec:conclusion}

PUMA separates geometric measurement, candidate selection and language-model verification. Auditing it on two pre-registered splits shows that calibrated rejection bounds the rate of gross errors but not their magnitude, and that the effective safeguard is an anatomically defined admissibility region rather than a tuned score threshold. The audit framework transfers directly to patient-level multi-centre data and to any measurand whose plausible range is known in advance.

\emph{Availability and data.} Stored predictions, weights and scripts support these analyses; new generation needs API access. HC18, FETAL\_PLANES\_DB and SLAKE are CC-BY-4.0, VQA-RAD CC0-1.0, PMC-VQA CC-BY-SA, Huatuo26M-Lite Apache-2.0; records with unresolved reuse terms are excluded. No new patient data were collected. \emph{Data use.} All data are existing public, de-identified archives; no ethics approval was required.

\bibliographystyle{splncs04}
\bibliography{main}

\end{document}